В.В. Кромер

## Обработка тестовых матриц, скорректированных на угадывание

При тестировании возникает проблема, связанная с попытками тестируемых угадать верный ответ. Эта проблема наиболее актуальна при тестировании заданиями закрытой формы с выбором одного правильного ответа из нескольких предложенных. Известно несколько подходов, связанных с обработкой тестовых результатов при наличии фактора угадывания.

Первый подход (наиболее простой) состоит в игнорировании самой проблемы, и реализован в широкомасштабной процедуре Единого государственного экзамена (ЕГЭ). Тестируемый поощряется (баллом) за удавшуюся попытку угадывания, неудавшаяся попытка никак не фиксируется и не влияет на результат. Подобный тест, как известно, обладает низкой дискриминантной валидностью, поскольку кроме намеченного для измерения свойства (уровня подготовленности по школьной химии или географии) измеряет максимальное из двух не предназначавшихся для измерения свойств личности – общей смышлености и специальной осведомленности. Получить нужную степень осведомленности тестируемый может у своего репетитора, который, в свою очередь, с вопросом наверняка знаком по статье [10]. Таким образом, в проигрыше оказываются вполне добросовестные тестируемые, учившие химию (географию) и не поднаторевшие в тонкостях тестирования. Отсутствие коррекции на догадку в результатах ЕГЭ называется одним из источников явных погрешностей измерения [2, с. 27].

Второй подход (самый жесткий) предполагает вычитание из суммы баллов, набранных тестируемым за правильные ответы, по одному баллу (штрафному) за каждый ошибочный ответ [1, с. 42]. Предполагается (явно или неявно), что ошибочный ответ – следствие неудавшейся попытки угадывания, т.е. нарушения инструкции, предписывающей выбирать ответ лишь в случае полной уверенности в его правильности. Здесь, как будет показано в дальнейшем, налицо несоразмерность строгости наказания тяжести проступка.

Третий подход (наиболее справедливый) заключается в коррекции тестового результата таким образом, чтобы математическое ожидание скорректированного тестового балла равнялось гипотетическому тестовому баллу, показанному тестируемым в условиях отказа от угадывания. Коррекция осуществляется по формуле $T_{i\,кор} = T_{i\,исх} - \frac{W_i}{m-1}$, где $T_{i\,кор}$ – скорректированный на угадывание тестовый балл $i$-го тестируемого, $T_{i\,исх}$ – исходный (сырой, первичный) тестовый балл, $W_i$ – кол-во ошибочных ответов, $m$ – кол-во предусмотренных вариантов ответа к каждому заданию.

В формуле предполагается, что *m* для всех тестовых заданий одинаково, что сужает ее возможности.

Легко видеть, что при $m = 2$ (случай т.н. заданий альтернативных ответов) третий и второй подходы совпадают. Именно в случае $m = 2$ вычитание одного балла за каждый ошибочный ответ вполне справедливо. При $m > 2$ наказание штрафным баллом за каждый ошибочный ответ слишком сурово.

Четвертый подход предложен нами [4] и заключается во внесении в матрицу тестовых результатов (традиционно заполняемую лишь нулями и единицами в зависимости от исхода "противоборства" тестируемого с заданием) отрицательных дробных величин $\left(-\dfrac{1}{m_j - 1}\right)$ за ошибочные ответы, где $m_j$ – количество вариантов ответа на *j*-е задание ($m_j$ может меняться для различных заданий).

Заполнение матрицы и последующий подсчет сумм (с учетом знаков) элементов матрицы по строкам и столбцам дает несмещенные значения, математическими ожиданиями которых являются гипотетические истинные баллы тестируемого и задания, что позволяет объективно оценить уровень подготовленности тестируемого и уровень трудности задания.

Как известно, тестируемый характеризуется, помимо уровня подготовленности, также степенью структурированности знаний, а задание, помимо степени трудности, еще и дифференцирующей способностью. Обе эти характеристики определяются совершенно одинаково (но по отношению каждая к своей стороне тестирования – тестируемому или заданию) и отражают «правильность» соответствующего профиля – последовательностей нулей и единиц соответствующей тестируемому строки и соответствующего заданию столбца. Идеальными (и практически нереализуемыми) являются профили с концентрацией всех единиц в левой части строки (или верхней части столбца) при условии, что матрица тестирования дважды упорядочена (по убыванию тестовых баллов тестируемых и баллов заданий).

Структурированность знаний тестируемого и дифференцирующая способность задания определяются исходя из меры отклонения фактического профиля от идеального (от степени «дырявости», т.е. чередования единиц и нулей в соответствующей строке или столбце). В рамках классической теории тестов дифференцирующую способность задания принято оценивать по коэффициенту корреляции Пирсона вектор-столбца задания с вектор-столбцом исходных тестовых баллов тестируемых. Поскольку вектор-столбец задания включен в вектор-столбец тестовых баллов тестируемых, оцениваемая дифференцирующая способность задания завышена, и это смещение тем больше, чем меньше в тесте заданий. Выходом является вычисление коэффициента корреляции вектор-столбца задания с вектор-столбцом тестовых баллов тестируемых за исключением баллов рассматриваемого задания. Вся процедура оценки дифференцирующей способности заданий легко реализуется посредством функции ПИРСОН программы MS Excel. Задания с коэффициентом корреляции $r \geq 0{,}2$ условно считаются пригодными (валидными) вплоть до более жесткой проверки по другим критериям.

Наличие явления угадывания вносит коррективы в описанную схему. Дадим определения некоторым понятиям, которыми мы будем пользоваться в дальнейшем. Истинной назовем матрицу, в которой угадывание не имеет места. Истинная матрица – гипотетическое теоретическое понятие, при обработке результатов сталкиваться с ней не приходится. Но именно характеристики сторон тестирования, извлеченные из истинной матрицы, являются несмещенными оценками истинных характеристик.

Искаженная матрица – матрица при реализации в процессе заполнения и обработки первого подхода. Характеристики сторон тестирования, извлекаемые из искаженной матрицы, как правило смещены.

Скорректированной матрицей назовем матрицу с внесенными в нее отрицательными корректирующими элементами при реализации четвертого подхода.

Целью данной статьи является разработка метода обработки скорректированных матриц, дающая характеристики сторон тестирования, идентичные характеристикам, извлекаемым из истинной матрицы.

При реализации первого подхода к проблеме угадывания часть нулей в истинной матрице заполняется единицами, обязанными своим происхождением исключительно успешному угадыванию. Далее такие единицы будем называть «угаданными» единицами. Вектор-столбцы заданий изменяются, и наибольшему искажению подвергаются вектор-столбцы трудных заданий, поскольку в них больше нулей, и они потенциально более подвержены искажениям.

Оценим характер изменения коэффициента корреляции задания с тестовыми результатами в искаженной матрице по сравнению с истинным коэффициентом корреляции. Искажение вектор-столбца задания не изменяет характера переменной, представленной в столбце – она остается дихотомической. Для частного случая, когда одна из переменных является дихотомической, коэффициент корреляции Пирсона вырождается в точечно-бисериальный коэффициент корреляции

$$r_{j\,pb} = \frac{M_{j1} - M_{j0}}{s_y} \sqrt{\frac{n_{j0} n_{j1}}{n(n-1)}}, \qquad (1)$$

где $M_{j1}$ – среднее арифметическое тестовых баллов тестируемых, получивших по $j$-му заданию 1 балл, $M_{j0}$ – среднее арифметическое тестовых баллов тестируемых, получивших по $j$-му заданию 0 баллов, $s_y$ – стандартное отклонение тестовых баллов, $n_{j0}$ – число тестируемых, получивших в задании 0 баллов, $n_{j1}$ – число тестируемых, получивших в задании 1 балл, $n = n_{j0} + n_{j1}$ – общее число тестируемых [1, с. 172].

При изменении вектор-столбцов заданий меняется и вектор-столбец тестовых результатов. Покажем, что переменная тестовых результатов подвергается при этом линейному преобразованию. Если $T_{i\,ист}$ – истинный тестовый балл $i$-го тестируемого, $k$ – максимально возможный тестовый балл тестируемого, равный количеству тестовых заданий, а $c$ – некий средневзвешенный по всем заданиям и тестируемым фактор, отражающий интенсивность и успешность попыток к угадыванию, т.е. среднюю долю нулей в истинной матрице, замещающихся единицами в искаженной матрице, математическое ожидание измеренного тестового балла тестируемого увеличится и составит $T_{i\,изм} = T_{i\,ист} + c(k - T_{i\,ист}) = (1-c)T_{i\,ист} + ck$. Видно, что истинный тестовый балл при угадывании подвергается линейному преобразованию, что позволяет (с целью упрощения анализа изменения $r_{j\,pb}$) считать столбец тестовых результатов неизменным. (Известно, что при линейном преобразовании одного или обоих рядов данных коэффициент корреляции Пирсона между ними не изменяется). В формуле (1) при обработке искаженной матрицы в соответствии с первым подходом $M_{j1}$ уменьшается (по 1 баллу за $j$-е задание получают в т.ч. и более слабые тестируемые с меньшим тестовым баллом), $M_{j0}$ в среднем не изменяется (нули заменяются единицами относительно равномерно по всему столбцу), $s_y$ остается неизменным при постоянстве столбца тестовых результатов, $n$ также неизменно. Остается проанализировать характер изменения $n_{j0}n_{j1} = n_{j0}(n - n_{j0})$. Для заданий с истинным $n_{j0} < 0{,}5n$ (т.е. заданий с уровнем трудности ниже среднего) значение производной $\dfrac{d}{d\,n_{j0}}[n_{j0}(n - n_{j0})] = n - 2n_{j0}$ положительно, т.е. $n_{j0}n_{j1}$ в искаженной матрице уменьшается по сравнению с истинным его значением за счет уменьшения $n_{j0}$, что в целом ведет к уменьшению $r_{j\,pb}$. При $n_{j0} > 0{,}5n$ (т.е. для заданий с уровнем трудности выше среднего) значение производной отрицательно, и угадывание ведет к увеличению произведения $n_{j0}n_{j1}$, что для очень трудных заданий не компенсируется уменьшением $M_{j1}$. В таких случаях возможно сохранение или незначительное увеличение $r_{j\,pb}$.

Рассмотрим характер изменения $r_{j\,pb}$ при четвертом подходе. Ряд значений элементов матрицы $x_{ij}$ в этом случае уже не является дихотомическим, и коэффициент корреляции Пирсона вычисляется по общей формуле [3, с. 360]:

$$r_j = \frac{\sum_{i=1}^{n}(x_{ij}-\bar{x}_j)(y_i-\bar{y})}{\sqrt{\sum_{i=1}^{n}(x_{ij}-\bar{x}_j)^2 \sum_{i=1}^{n}(y_i-\bar{y})^2}}, \qquad (2)$$

где $y_i = \sum_{j=1}^{k} x_{ij}$ – тестовый балл $i$-го тестируемого, $\bar{x}_j = \frac{1}{n}\sum_{i=1}^{n} x_{ij}$ – среднее значение балла задания, а $\bar{y} = \frac{1}{n}\sum_{i=1}^{n} y_i$ – среднее значение тестовых баллов по всем тестируемым.

Преобразуем (2) к виду

$$r_j = \frac{\sum_{i=1}^{n} x_{ij} y_i - n\bar{x}_j \bar{y}}{\sqrt{\sum_{i=1}^{n} x_{ij}^2 - n\bar{x}_j^2}\sqrt{\sum_{i=1}^{n} y_i^2 - n\bar{y}^2}} \qquad (3)$$

на основе формул из [3, с. 100, 361].

При внесении в истинную матрицу корректирующих отрицательных элементов в каждом столбце задания на каждый угаданный единичный балл в среднем придется $(m_j - 1)$ корректирующих элементов $\left(-\frac{1}{m_j - 1}\right)$, что не изменяет средних значений $\bar{x}_j$ и $\bar{y}$. Корректирующие элементы располагаются в столбце задания вблизи элемента с угаданным баллом, что при достаточно большом $n$ обеспечивает относительное постоянство всех значений $y_i$ на данном участке столбца. Таким образом, в скорректированной матрице при расчете по формуле (3) не меняется значение суммы $\sum_{i=1}^{n} x_{ij} y_i$, а соответственно и значение числителя выражения. В знаменателе формулы (3) не меняется также значение $\sqrt{\sum_{i=1}^{n} y_i^2 - n\bar{y}^2}$, поскольку в среднем не изменяются тестовые результаты тестируемых (компенсация угаданных баллов осуществляется не только по столбцам, но и по строкам матрицы).

Изменяется (меняется в бо́льшую сторону) лишь значение $\sqrt{\sum_{i=1}^{n} x_{ij}^2 - n\bar{x}_j^2}$, поскольку ряд нулей в столбце замещается элементами, равными отрицательным значениям и единицам, что при возведении их в квадрат увеличивает сумму $\sum_{i=1}^{n} x_{ij}^2$. Таким образом, расчет коэффициента корреляции по данным скорректированной матрицы по формуле (3) дает заниженное по сравнению с истинным значение.

Встает задача выведения некоего конструкта на основе элементов столбца задания и столбца тестовых результатов скорректированной тестовой матрицы, значение которого будет в среднем равно коэффициенту корреляции Пирсона, рассчитанному по истинной матрице. Источником смещения значения коэффициента корреляции является сумма $\sum_{i=1}^{n} x_{ij}^2$, искажаемая ввиду добавления в столбец $j$-го задания угаданных единичных баллов и корректирующих элементов $\left(-\dfrac{1}{m_j - 1}\right)$. Каждый угаданный единичный балл формально не отличается от истинного (добытого на основе знания) и при возведении в квадрат дает 1. Для компенсации искажения достаточно прибавить к сумме $\sum_{i=1}^{n} x_{ij}^2$ сумму всех значений корректирующих элементов по данному столбцу $\sum_{i=1}^{n}\left(-\dfrac{1}{m_j - 1}\right)$. Поскольку для нулей и единиц квадрат равен самому значению, сумма $\sum_{i=1}^{n} x_{ij}^2 + \sum_{i=1}^{n}\left(-\dfrac{1}{m_j - 1}\right)$ в среднем равна сумме $\sum_{i=1}^{n} x_{ij}$.

Таким образом, истинный коэффициент корреляции вектор-столбца задания с вектор-столбцом тестовых баллов тестируемых следует считать по формуле

$$r_{j\,ucm} = \dfrac{\sum_{i=1}^{n} x_{ij} y_i - n\,\overline{x}_j\,\overline{y}}{\sqrt{\sum_{i=1}^{n} x_{ij} - n\,\overline{x}_j^2}\sqrt{\sum_{i=1}^{n} y_i^2 - n\,\overline{y}^2}}. \qquad (4)$$

Чтобы избежать вычислений по формуле (4), разделим (4) на (3) для получения поправочного коэффициента $K_j$, что позволит воспользоваться для вычисления истинного коэффициента корреляции стандартными функциями программы MS Excel:

$$K_j = \frac{r_{j\,ucm}}{r_j} = \sqrt{\frac{\sum_{i=1}^{n} x_{ij}^2 - n\bar{x}_j^2}{\sum_{i=1}^{n} x_{ij} - n\bar{x}_j^2}} = \sqrt{\frac{\sum_{i=1}^{n}(x_{ij} - \bar{x}_j)^2}{n(\bar{x}_j - \bar{x}_j^2)}}. \qquad (5)$$

Поскольку выборочное значение дисперсии ряда значений $x_{ij}$ равно $s_{j\,выб}^2 = \frac{\sum_{i=1}^{n}(x_{ij} - \bar{x}_j)^2}{n-1}$ [3, с. 99], можно записать $K_j = \sqrt{\frac{(n-1)\,s_{j\,выб}^2}{n(\bar{x}_j - \bar{x}_j^2)}}$. Существует также понятие дисперсии по генеральной совокупности $s_{j\,ген}^2 = \frac{\sum_{i=1}^{n}(x_{ij} - \bar{x}_j)^2}{n}$, равное выборочной дисперсии, умноженной на $\frac{n-1}{n}$. Среднее значение балла задания $\bar{x}_j$ в тестологии обозначается за $p_j$ и по определению равно доле испытуемых, давших правильные ответы на $j$-е задание (при подсчете $p_j$ по искаженной матрице невольно учитываются и угаданные баллы). В скорректированной матрице $p_j$ и $\bar{x}_j$ согласно определению не совпадают, поскольку доля единичных элементов в столбце задания выше среднего значения элементов столбца, однако при формальном подсчете $p_j$ как среднего значения элементов столбца, значения $p_j$ и $\bar{x}_j$ совпадают. Окончательно формула для $K_j$, пригодная для расчетов по программе MS Excel, приобретает вид $K_j = \sqrt{\frac{s_{j\,ген}^2}{p_j(1-p_j)}}$. Значение $p_j$ вычисляется по функции СРЗНАЧ, значение $s_{j\,ген}^2$ – по функции ДИСПР. Истинные значения коэффициента корреляции Пирсона, отражающие валидность задания, находятся по формуле $r_{j\,ucm} = K_j r_j$, где $r_j$ – коэффициент корреляции, найденный по скорректированной матрице по функции ПИРСОН.

При отборе тестовых заданий иногда возникает необходимость оценить интеркорреляции – коэффициенты корреляции Пирсона каждого задания с каждым. Коэффициенты корреляции между заданиями $s$ и $t$ в общем случае вычисляются по формуле, справедливой для истинной матрицы:

$$r_{st} = \frac{\sum_{i=1}^{n} x_{is} x_{it} - n\bar{x}_s \bar{x}_t}{\sqrt{\sum_{i=1}^{n} x_{is}^2 - n\bar{x}_s^2}\sqrt{\sum_{i=1}^{n} x_{it}^2 - n\bar{x}_t^2}}. \qquad (6)$$

При добавлении в истинную матрицу угаданных баллов и корректирующих баллов (т.е. преобразовании ее в скорректированную матрицу) значение числителя в формуле (6) не меняется, а характер изменения выражений типа $\sqrt{\sum_{i=1}^{n} x_{ij}^2 - n\bar{x}_j^2}$ уже нами исследован (подобные выражения увеличиваются в $K_j$ раз, где вместо $j$ подставляются $s$ или $t$). Таким образом, коэффициент корреляции Пирсона между векторами-столбцами заданий $s$ и $t$, вычисленный по скорректированной матрице, уменьшается в $K_s K_t$ раз по сравнению с истинным коэффициентом.

Общая технология работы со скорректированной матрицей следующая: для каждого $j$-го задания вычисляется корректирующий коэффициент $K_j = \sqrt{\dfrac{s_{j\,ген}^2}{p_j(1-p_j)}}$. После определения коэффициента корреляции Пирсона задания с тестовыми баллами тестируемых полученное значение необходимо умножить на соответствующее заданию значение $K_j$. При оценке коэффициента интеркорреляции определенное значение коэффициента корреляции между заданиями $s$ и $t$ необходимо умножить на $K_s K_t$. В каждом из этих случаев в результате получается несмещенная оценка коэффициента корреляции или интеркорреляции для истинной матрицы.

Рассмотрим обработку скорректированной тестовой матрицы для оценки коэффициента надежности теста. Определение коэффициента надежности методом повторного тестирования, когда за надежность теста принимается коэффициент корреляции Пирсона между результатами двух разнесенных во времени тестирований одной и той же выборки одним и тем же тестом, искажений в случае использования скорректированных матриц не вносит, поскольку тестовые результаты в скорректированной матрице являются математическим ожиданием результатов истинной матрицы.

Сказанное справедливо и при оценке надежности методом расщепления теста на две половинки как по коэффициенту корреляции между двумя половинками, так и при оценке надежности полутеста по формуле Рюлона $r_{нт} = 1 - \dfrac{s_d^2}{s_y^2}$, где $s_d$ – стандартное отклонение разностей между результатами каждого тестируемого по обеим половинкам теста, $s_y$ – стандартное отклонение тестовых баллов тестируемых [8, с. 196–197]. (Интересующая нас надежность полного теста в данном случае должна быть определена по известной формуле Спирмена–Брауна) [8, с. 197].

Несмещенную оценку надежности дает также известная формула Кьюдера–Ричардсона KR-20: $r_{нm} = \dfrac{k}{k-1}\left(1 - \dfrac{\sum_{j=1}^{k} p_j(1-p_j)}{s_y^2}\right)$, где $k$ – число тестовых заданий, $s_y$ – стандартное отклонение тестовых баллов тестируемых, а $p_j$ – среднее значение элементов столбца $j$-го задания [1, с. 215]. При расчете надежности по истинной матрице $p_j$ полагается равным доле тестируемых, ответивших верно на $j$-е задание. Поскольку среднее значение балла задания в скорректированной матрице равно доле успешно справившихся с заданием тестируемых по истинной матрице, формула пригодна для оценки надежности.

Однако формула «альфа Кронбаха» $\alpha = \dfrac{k}{k-1}\left(1 - \dfrac{\sum_{j=1}^{k} s_j^2}{s_y^2}\right)$, где $s_j$ – стандартное отклонение значений элементов вектор-столбца $j$-го задания [7, с. 207], непригодна для расчета надежности, поскольку в скорректированной матрице значение $s_j^2 > p_j(1-p_j)$.

Рассмотрим возможности обработки скорректированных матриц в рамках современных моделей тестирования. К современным моделям тестирования относятся модели различной сложности, связывающие вероятность успеха тестируемого со свойствами задания и самого тестируемого.

В основу функции успеха кладется какая-либо из известных функций распределения (поскольку именно функции распределения отвечают требованиям, предъявляемым к функции успеха). В теоретических разработках используется нормальная функция распределения, на практике заменяемая близкой к ней логистической, исходя из удобства ее вычисления (выражается через элементарные функции).

Методологически наиболее удобно рассмотреть функцию успеха современных моделей тестирования как частный случай наиболее общей (и последней по времени публикации) формулы функции успеха из [5]:

$$P_{ij} = c_i c_j + (1 - c_i c_j)\left[1 + \exp\left(\dfrac{d_i d_j}{\sqrt{d_i^2 d_j^2}}(\delta_j - \theta_i)\right)\right]^{-1}, \qquad (7)$$

где $\theta_i$ – потенциал тестируемого, $\delta_j$ – потенциал задания, $d_i$ и $d_j$ – соответственно избирательность тестируемого и задания, а $c_i$ и $c_j$ – соответственно готовность тестируемого к угадыванию и свойство задания, провоцирующее угадывание. Потенциалом тестируемого и задания мы называем соответственно уровень подготовленности тестируемого и уровень трудности задания, а избирательностью тестируемого и задания – соответственно степень структурированности знаний тестируемого и дифференцирующую способность задания.

При $c_i = c_j = 0$, $d_i = d_j = \sqrt{2}$ мы получаем однопараметрическую модель Раша $P_{ij} = [1 + \exp(\delta_j - \theta_i)]^{-1}$, при $c_i = c_j = 0$, $d_i = +\infty$ – двухпараметрическую модель Бирнбаума $P_{ij} = [1 + \exp(d_j(\delta_j - \theta_i))]^{-1}$, при $c_i = 1$, $d_i = +\infty$ – трехпараметрическую модель Бирнбаума $P_{ij} = c_j + (1 - c_j)[1 + \exp(d_j(\delta_j - \theta_i))]^{-1}$. Менее известна (и практически не применяется) двухпараметрическая модель Бирнбаума в форме $P_{ij} = [1 + \exp(d_i(\delta_j - \theta_i))]^{-1}$ (в наиболее общей формуле (7) полагается $c_i = c_j = 0$, $d_j = +\infty$). Возможны также и иные частные функции успеха при соответствующем выборе параметров $c_i$, $c_j$, $d_i$ и $d_j$.

Формула (7) является пятипараметрической (в соответствии с общепринятым порядком считать разность потенциалов задания и тестируемого $(\delta_j - \theta_i)$ одним параметром), и численное решение ее относительно пяти параметров связано с возможностью возникновения ложных решений на основе мультимодальности функции правдоподобия. Ниже мы покажем, что необходимость в пятипараметрической модели возникает лишь при обработке искаженной матрицы данных. При обработке скорректированной на угадывание матрицы необходимость в параметрах $c_i$ и $c_j$, связанных с вероятностью угадывания, исчезает, и пятипараметрическая функция успеха (7) вырождается до трехпараметрической функции вида

$$P_{ij} = \left[1 + \exp\left(\frac{d_i d_j}{\sqrt{d_i^2 d_j^2}}(\delta_j - \theta_i)\right)\right]^{-1}, \qquad (8)$$

численная параметризация которой более устойчива, чем функции (7) [6].

Для определения характеристик сторон тестирования, как известно, используется метод наибольшего правдоподобия Фишера. При дихотомических данных логарифмическая функция правдоподобия для вектора $\vec{x}_i = \{x_{i1}, x_{i2}, ... x_{ik}\}$ записывается в виде $\ln L_i\{\vec{x}_i | \theta_i\} = \sum_{j=1}^{k}\{x_{ij} \ln P_{ij} + (1 - x_{ij})\ln(1 - P_{ij})\}$ [9, с. 22–24], а для всей матрицы общая функция правдоподобия имеет вид:

$$\ln L_i = \sum_{i=1}^{n}\sum_{j=1}^{k}\{x_{ij} \ln P_{ij} + (1 - x_{ij})\ln(1 - P_{ij})\}. \qquad (9)$$

Покажем, что формула (9) может быть обобщена на случай скорректированной тестовой матрицы с нулевыми, единичными и отрицательными элементами.

Вероятность дихотомического элемента $x_{ij}$ составляет
$$p = P_{ij}^{x_{ij}} (1-P_{ij})^{(1-x_{ij})}, \qquad (10)$$
что следует из последовательной подстановки $x_{ij}=1$ и $x_{ij}=0$, что дает соответственно $p=P_{ij}$ и $p=1-P_{ij}$. При отрицательных значениях $-1<x_{ij}<0$ формально вычисляемые по формуле (10) вероятности
$$p = P_{ij}^{-|x_{ij}|}(1-P_{ij})^{1+|x_{ij}|} = (1-P_{ij})\left(\frac{1-P_{ij}}{P_{ij}}\right)^{|x_{ij}|}$$
при достаточно малых $P_{ij}$ превышают 1, т.е. являются не имеющими смысла. В то же время в рамках принятой модели предполагается, что каждые $(m-1)$ элементов $\left(-\frac{1}{m-1}\right)$ уравновешены одним угаданным баллом 1. Сумма рассматриваемых $m$ элементов дает 0, что позволяет считать, что произведение вероятностей рассматриваемых $m$ элементов равно вероятности $m$ нулевых элементов, т.е. $(1-P_{ij})^m$. В формуле учтено значение $P_{ij}$ для элемента $x_{ij}$, хотя рассматриваемые $m$ элементов линейно разнесены в строке (столбце). Однако рассмотрение модели с достаточно большим $n$ (или $k$) позволяет считать $P_{ij}$ общим для всех $m$ элементов.

Приравняем вероятность рассматриваемых $m$ элементов вероятности $m$ нулевых элементов, обозначив за $P_x$ неизвестную искомую вероятность корректирующего элемента $x_{кор}$: $P_x^{m-1} P_{ij} = (1-P_{ij})^m$, откуда
$$P_x^{m-1} = \frac{(1-P_{ij})^m}{P_{ij}} \text{ и } P_x = \frac{(1-P_{ij})^{\frac{m}{m-1}}}{P_{ij}^{\frac{1}{m-1}}}.$$
Поскольку $x_{кор}=-\frac{1}{m-1}$, окончательно получаем $P_x = P_{ij}^{x_{кор}}(1-P_{ij})^{1-x_{кор}}$. Таким образом, параметризация модели по функции успеха (8) осуществляется общепринятым методом с подсчетом логарифмической функции правдоподобия по формуле (9), обсчитывая единообразно нули, единицы и отрицательные корректирующие элементы.

Специфичной особенностью обработки скорректированной тестовой матрицы является необходимость предварительного удаления из нее, помимо строк и столбцов, состоящих целиком из нулей и единиц (что общепринято), также строк и столбцов с отрицательной общей суммой элементов. Причиной возникновения подобных строк является устойчивое отрицательное знание тестируемых, что выводит тестируемых за рамки оценки уровня их подготовленности тестовыми моделями (рассчитанными на оценку положительных знаний). Столбцы с отрицательной суммой элементов характерны для заданий повышенной трудности с неравной привлекательностью правильного ответа и дистракторов для незнающего тестируемого. В сбалансированном же по ответной части задании вероятность правильного ответа является функцией потенциала тестируемого и всегда превышает вероятность выбора дистрактора, стремясь к ней при неограниченном уменьшении потенциала тестируемого.

## Список литературы